\documentclass[10pt,twocolumn,letterpaper]{article}

\usepackage{cvpr}
\usepackage{times}
\usepackage{epsfig}
\usepackage{graphicx}
\usepackage{amsmath}
\usepackage{amssymb}
\usepackage{subcaption}
\usepackage{fixltx2e}
\usepackage{multirow}
\usepackage{comment}

\DeclareMathOperator{\E}{\mathbb{E}}
\usepackage[pagebackref=true,breaklinks=true,letterpaper=true,colorlinks,bookmarks=false]{hyperref}

\cvprfinalcopy % *** Uncomment this line for the final submission

\def\cvprPaperID{3359} % *** Enter the CVPR Paper ID here

\ifcvprfinal\fi
\begin{document}

%%%%%%%%% TITLE
\title{GazeGAN -- Unpaired Adversarial Image Generation for Gaze Estimation}

\author{Matan Sela\thanks{Work was done when author was at Google Research.}\\
{\tt\small matanasel@gmail.com}\\
Technion, Israel
\and
Pingmei Xu\\
{\tt\small pingmeix@google.com}\\
Google Inc.
\and
Junfeng He\\
{\tt\small junfenghe@google.com}\\
Google Inc.
\and
Vidhya Navalpakkam\\
{\tt\small vidhyan@google.com}\\
Google Inc.
\and
Dmitry Lagun\\
{\tt\small dlagun@google.com}\\
Google Inc.
}
\maketitle
\pagestyle{empty}

%%%%%%%%% ABSTRACT
\begin{abstract}
Recent research has demonstrated the ability to estimate gaze on mobile devices by performing inference on the image from the phone's front-facing camera, and without requiring specialized hardware. While this offers wide potential applications such as in human-computer interaction, medical diagnosis and accessibility (e.g., hands free gaze as input for patients with motor disorders), current methods are limited as they rely on collecting data from real users, which is a tedious and expensive process that is hard to scale across devices. There have been some attempts to synthesize eye region data using 3D models that can simulate various head poses and camera settings, however these lack in realism.

In this paper, we improve upon a recently suggested method, and propose a generative adversarial framework to generate a large dataset of high resolution colorful images with high diversity (e.g., in subjects, head pose, camera settings) and realism, while simultaneously preserving the accuracy of gaze labels. The proposed approach operates on extended regions of the eye, and even completes missing parts of the image. Using this rich synthesized dataset, and without using any additional training data from real users, we demonstrate improvements over state-of-the-art for estimating 2D gaze position on mobile devices. %Our approach works out-of-the-box and provides robust gaze estimation even when real labelled data is limited.
We further demonstrate cross-device generalization of model performance, as well as improved robustness to diverse head pose, blur and distance.

\end{abstract}

%%%%%%%%% BODY TEXT
\section{Introduction}

\begin{figure}[t]
\begin{center}
%\fbox{\rule{0pt}{3in} \rule{1.0\linewidth}{0pt}}
   \includegraphics[width=1.0\linewidth]{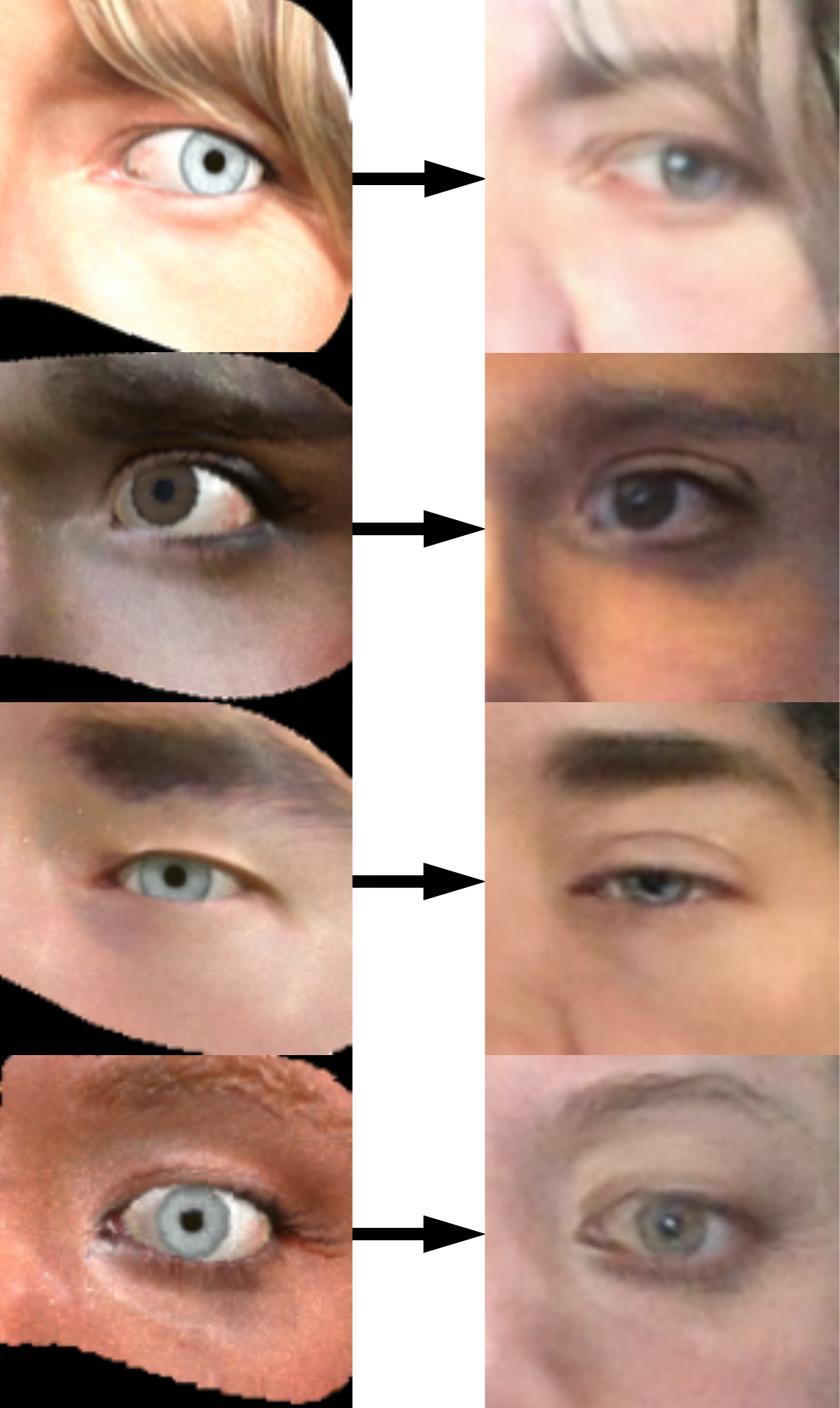}

\vspace{-15pt}
\end{center}
   \caption{Synthetic images (left) alongside their refined realistic counterparts, generated by our framework (right).}
\vspace{-20pt}
\label{fig:teaser}
\end{figure}
Automated estimation of gaze direction is a fundamental component in numerous applications.
Human-computer interaction~\cite{jacob2003eye, majaranta2014eye, morimoto2005eye, mutlu2009footing},
behavior monitoring~\cite{morimoto2005eye, bulling2011eye}, vision-systems~\cite{papadopoulos2014training, sattar2015prediction}, AR/VR,
medical diagnoses~\cite{holzman1974eye}, and gaming~\cite{corcoran2012real} are just a few cases where the ability to track the eye is essential. A brief glance at the physiology of the human eye shows that it has a structure of a ball with a bulge, and a surprisingly uniform diameter of about 24mm for adults. The transparent bump, called cornea, covers the iris, and is approximately 11mm wide.
Due to the uniform scale and the geometric structure of the eye, it is possible to detect the gaze direction from a well lit image of entirely visible eye just by analyzing the iris contour~\cite{wood2015rendering}. However, in common setups, the iris contour is mostly occluded by the eyelid and the eyelash.

Alternative solutions for handling this occlusion have been proposed~\cite{chen20083d, wood2014eyetab, zhang2015appearance, huang2015tabletgaze, wood2015rendering, funes2016gaze}. With the recent consistent success of deep models in performing computer vision tasks, CNN-based approaches have become popular~\cite{baluja1994non, zhang2015appearance, krafka2016eye, zhang2016s}. The ability to provide accurate predictions for unseen general cases is a key advantage of deep networks over shallow architectures. However, the performance is highly dependent upon the availability of a large set of annotated training images. Ideally, the training examples are acquired from the real world. Yet, for an abundance of computer vision problems, including gaze estimation, collecting a dataset with large diversity from real users is a tedious and expensive process.

In a recent paper, Shrivastava et al.~\cite{shrivastava2017learning} tried to address this problem by suggesting a framework for synthesizing realistic images of the eye region. They employed a generative adversarial network model, where instead of a randomized vector, the input to the network is a synthetic image. To preserve the annotation of synthetic examples under the domain change, they penalized the pixel level deviation between simulated images and their refined versions. However, this approach has several limitations. Due to the mode collapse phenomenon which restricts the expressiveness of distribution functions in generative adversarial networks, the diversity of the resulting images is limited. In addition, the method could only process small gray scale images (36x60) due to the unstable training of adversarial networks. Further, as shown in~\cite{isola2017image}, such an image-to-image translation framework performs better when pixel level alignment between image pairs is given in the training set, which is difficult to acquire. In contrast, our methodology treats the problem as an unpaired image-to-image translation one~\cite{zhu2017unpaired}.

In this paper, we address the above limitations by introducing a novel interplay between synthetic and real, unlabeled data. The proposed framework involves a simulator of the eye region~\cite{wood2015rendering} and a large and diverse set of real, unlabeled facial images.
Given a set of parameters representing pose, illumination conditions and gaze direction, the simulator renders images accordingly. To increase the diversity of the synthetic model, we automatically extract realistic textures of skin tissues around the eye from unlabeled facial images. Next, we simulate a large set of images of the eye region, for which the gaze direction is known and is drawn from a distribution which imitates the real world. To introduce realism, we propose a novel pixel level domain adaptation method, inspired by the recent unpaired image-to-image translation method of Park et al.~\cite{zhu2017unpaired}. The original framework employs two adversarial networks -- one which maps synthetic images to the realistic domain, the other one which learns the inverse direction -- and optimizes pixel level constraints on the maps, which take an image from one domain to the other and back to the original one. Here, we extend this approach for the purpose of gaze direction preservation. To this end, we pretrain a network to predict the gaze direction from synthetic images only, and use this network as a constraint on one of the cyclic maps.

The resulting framework generates significantly higher resolution color images with realism, while preserving the gaze direction annotation. Surprisingly, the method also completes missing domains in the synthetic images (see Figure\ref{fig:teaser}). Our contributions are listed below:
\begin{enumerate}
%\item An improved framework which fuses simulated (synthetic) and real, unlabelled data into a reliable annotated training data source. This framework is general and can be applied to numerous other computer vision tasks which are beyond the scope of this paper.
%\item An improved eye region simulator which can render unlimited amounts of skin textures from unlabeled facial images.
\item A framework which fuses simulated (synthetic) and real, unlabeled data into a reliable annotated training data source. We use this to generate high resolution, colorful, diverse and realistic looking images of the extended eye region, with precise gaze direction annotation.
\item An improvement over state-of-the-art in estimation of gaze position (2D) on mobile devices, without collecting any additional training data from real users.
\item Cross-device generalization of performance, and robust gaze estimation for diverse head pose and blur.
\end{enumerate}

\begin{figure*}
\begin{center}
%\fbox{\rule{0pt}{2in} \rule{1.0\linewidth}{0pt}}

\includegraphics[width=1.0\linewidth]{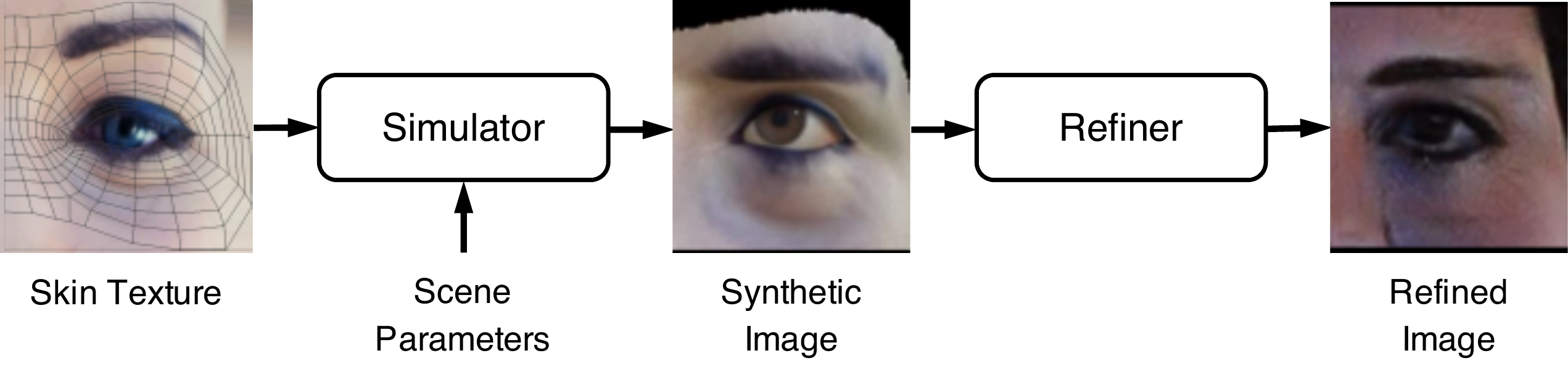}
\vspace{-30pt}
\end{center}
   \caption{Proposed image generation framework.}

\vspace{-10pt}
\label{fig:pipeline}
\end{figure*}

%-------------------------------------------------------------------------
\section{Related Work}
The problem of gaze direction estimation has recently become an active research area.
It is common to distinguish between two types of tasks -- 2D and 3D gaze estimation.
In 3D gaze estimation, the goal is to recover a unit vector representing the gaze direction, while in 2D gaze estimation the goal is to predict a point on a plane (usually a screen) posed in the scene where the gaze is directed towards.
While synthesizing data is easier for the former, it is simpler to acquire real data for the latter.

Traditionally, existing approaches for performing gaze estimation are classified as either model-based or appearance-based.
Appearance-based estimators are trained to regress the gaze direction directly from the image, while model-based methods try to model the geometric structure of the eye region.
A recent line of research suggests synthesizing images of the eye region for training appearance-based models. We discuss the different approaches below.

\textbf{Model-based gaze estimation:}
Several model-based gaze estimation methods use an infrared light source or high quality image sensor~\cite{guestrin2006general, nakazawa2012point, chen2011probabilistic}. This technique allows better separation of the iris from the rest of the image.
However, the additional hardware limits the scalability of these methods.
To simplify the process, other approaches try to fit a model to the entire face as a preprocessing step~\cite{ishikawa2004gaze, chen20083d}, requiring that the entire face is visible.
For scenarios in which major part of the face is occluded, Wood et al.~\cite{wood2014eyetab} proposed to fit a parametric model to the projection of the iris contour on the image plane.
Similar to this method, we decouple the problem of facial pose estimation from the problem of gaze estimation.
Thus, we analyze and synthesize images of the eye region only.

\textbf{Appearance-based gaze estimation:}
Appearance-based approaches estimate the gaze direction using a regression procedure based on eye images.
Due to lack of training data, a common simplification is to introduce additional knowledge, such as the head pose into the regression framework, or to train only shallow models~\cite{lu2014learning, schneider2014manifold,smith2013uist} or personalized models~\cite{zhang2015appearance}.
However, recently, Krafka et al.~\cite{krafka2016eye} have shown that given enough training data, a deep network can generalize prediction across subjects.
For doing so, they acquired the largest gaze estimation dataset currently available with almost fifteen hundred subjects and 1.5 million images.
They also showed that by increasing the number of identities in the dataset, the estimation error decreases significantly.
In this paper, we improve upon this approach by synthesizing data with unlimited cross-subject diversity.

\textbf{Training Image Synthesis:}
The idea of rendering and synthesizing training images for training appearance-based gaze estimation model was introduced by Yusuke et al. in~\cite{sugano2014learning}.
They proposed to recover the geometric structure of the face based on images taken from eight calibrated cameras.
To increase the dataset size they rendered images from unseen views of the textured facial geometry.
To enrich the diversity of the synthesized data, Wood et al.~\cite{wood2016learning} constructed a 3D morphable model which can generate different shapes and textures of the eye region.
This approach allows rendering images in an unlimited set of poses and lighting conditions while controlling the gaze direction of the 3D model.
Yet, the simulated images still look synthetic and the diversity is limited to linear combinations of a few tens of subjects.
Perhaps the most similar method to ours was proposed by Shrivastava et al.~\cite{shrivastava2017learning}, who used an adversarial network for mapping from the synthetic domain into the realistic one. As mentioned in the previous section and detailed in section 3, we make a number of improvements over this approach to increase the realism, diversity and accuracy of the generated dataset.

\section{Unpaired Training Image Synthesis}
\subsection{Overview}
%We propose to synthesize a large dataset of eye region images with a focus on diversity, realism and accurate gaze direction annotation.
The image synthesis pipeline is described in Figure \ref{fig:pipeline}.
%We start from a simulator which allows forming 3D scene of the eye region with specified lighting conditions, gaze direction, and skin shape.
A simulator forms a 3D scene of the eye region with specified lighting conditions, gaze direction, and skin shape.
To extend the limited diversity of the original principal component-based texture model, we automatically align facial images with the UV texture space of the 3D model.
This enables rendering image of the eye region with unlimited amount of textures. To improve the realism, an unpaired pixel level domain adaptation technique maps synthetic images to the realistic domain. This step requires the availability of real eye region images which are unlabeled.
Since the data is simulated, we can use the annotation of the gaze direction and pretrain a gaze direction estimator with the synthetic data.
Finally, to enforce the refiner network to preserve the gaze throughout the mapping, we use the pre-trained network as a constraint on the cycle of translation from synthetic to real and back to synthetic.

\subsection{Diversity}
\begin{figure}[t]
\begin{center}
%\fbox{\rule{0pt}{4in} \rule{1.0\linewidth}{0pt}}
   \includegraphics[width=0.8\linewidth]{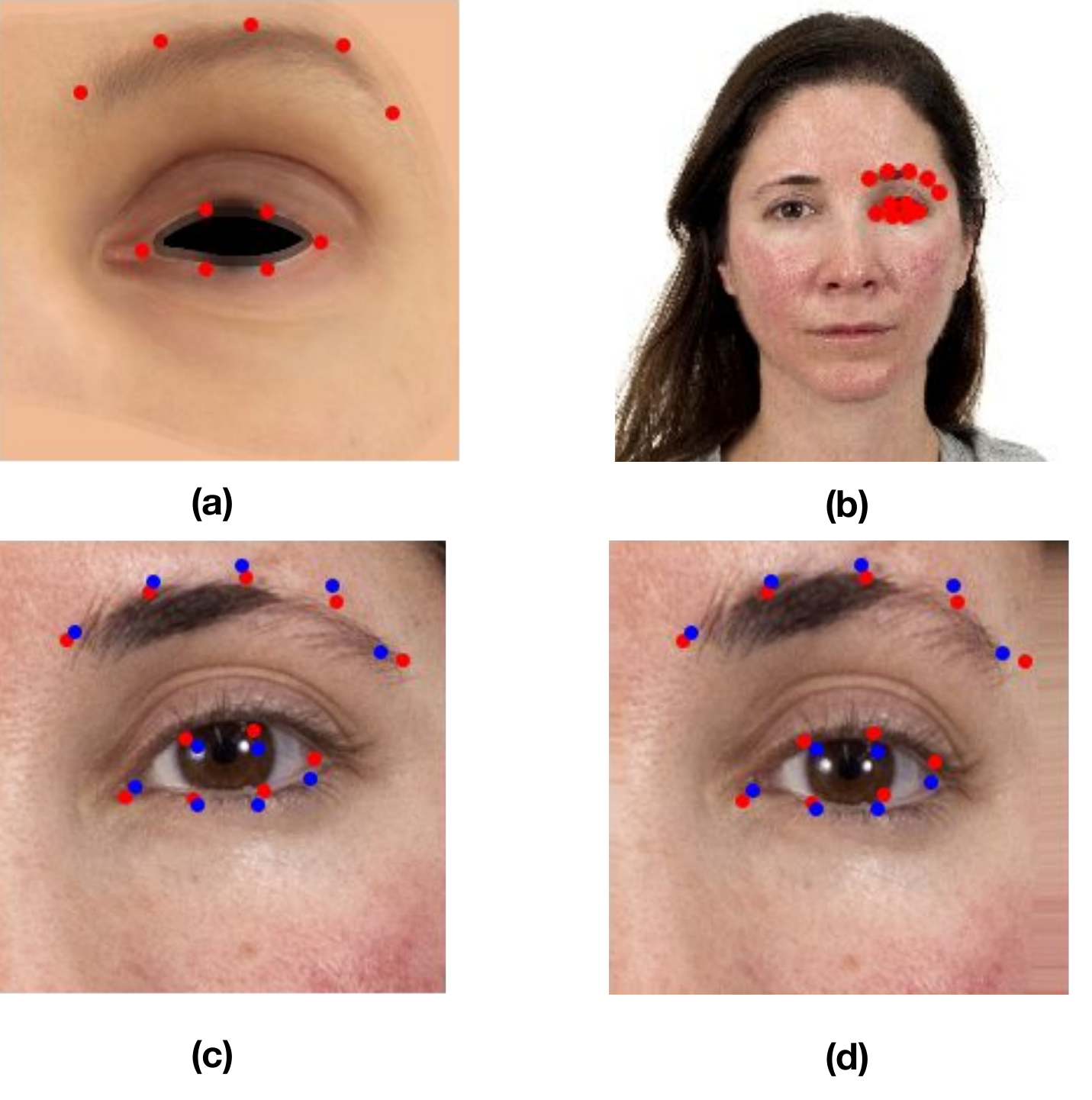}
\vspace{-20pt}
\end{center}
   \caption{Location of the eye region landmarks in the canonical UV texture (a). To extract the new texture: we automatically detect eye region landmarks in the real image (b); then we compute the optimal 2D similarity transform which aligns the facial image with the UV space of our 3D model \cite{wood2016learning} -- the result of this mapping shown at (c), the red circles correspond to the location of the landmarks in the canonical UV texture (a), blue circles correspond to the location of rigidly mapped landmarks; lastly, we perform a non rigid image warping to compute a pixel-wise alignment shown on (d). The example facial image was taken from the Chicago Face Database~\cite{ma2015chicago}. }
\vspace{-10pt}
\label{fig:texture_map}
\end{figure}

Our geometric 3D eye region surface is textured by a UV mapping technique~\cite{heckbert1986survey}. To increase the diversity of the skin texture, we use a large dataset of high resolution facial images taken from the front. Each image produces a single UV texture for the 3D model. The procedure is visualized in Figure \ref{fig:texture_map}. First, we detect landmark points on the facial images. Next, we compute the rotation, translation and scale which minimizes the Euclidean distance between the landmarks on the original facial image and the corresponding points on the UV space of the 3D model ~\cite{wood2016learning}. Finally, to account for non rigid deformation, we apply a smooth image warping based on the discrepancy between the source and target points~\cite{arad1994image}. An improved method to compute this pixel to pixel correspondence would be to modify the positions of the vertices using the analysis-by-synthesis method of Wood et al.~\cite{wood2017gazedirector}. However, for our purposes, we found that a simple similarity transformation followed by an image warping step is sufficient. With this procedure, we generated five million different skin textures. For fitting the distribution of the synthesized images to the real world, we adjusted the distribution of the simulated parameters to the distribution of the real dataset of Krafka et al.~\cite{krafka2016eye}.

%\vspace{-20pt}
\subsection{Realism} \label{sec:realism}
\begin{figure*}
\begin{center}
%\fbox{\rule{0pt}{2in} \rule{1.0\linewidth}{0pt}}
   \includegraphics[width=1.0\linewidth]{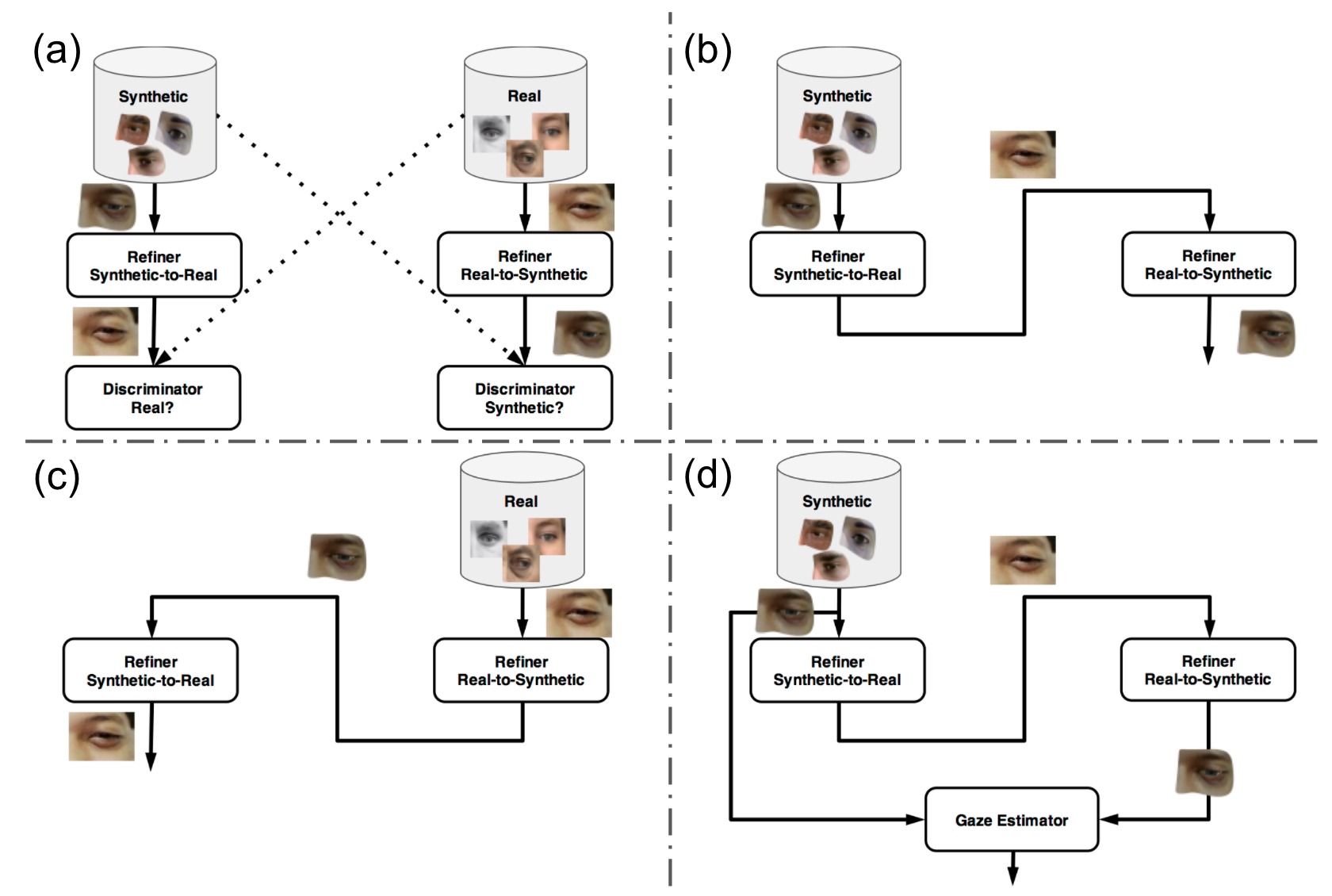}
\end{center}
\vspace{-25pt}
   \caption{Different components of the proposed loss function: (a) two independent generative adversarial networks (see Section \ref{sec:realism}); (b) synthetic-real-synthetic cycle consistency loss (see Section \ref{sec:realism}); (c) real-synthetic-real cycle consistency loss (see Section \ref{sec:realism}); (d) synthetic-real-synthetic gaze cycle consistency loss (see Section \ref{sec:gaze_preserve}).}
\vspace{-10pt}
\label{fig:refine_pipeline}
\end{figure*}

Although the texture produced by the previous step looks realistic on the skin region of the image, the eye itself still looks synthetic. Since the iris pose is perhaps the most important feature for detecting the gaze direction, for improving the performance, we need to generate data that looks realistic along this region as well.
For transferring the domain, we use a large set of unlabeled real images of the eye region taken from the Internet.
Since we do not have correspondence or alignment between our images from the two domains, we begin from the unpaired image-to-image translation method of Zhu et al.~\cite{zhu2017unpaired}.
This method maps images from one domain into another, where no pixel-wise alignment in the training data is given.
For example, the method was shown to map images of horses to zebras, or apples to oranges.
One of the admitted limitations of this approach is the inability to modify the geometric structure of the scene in the image.
We perceive this limitation as one of our features, as our goal is to preserve the position of the iris and the eyeball as well as the pose of the eye region in the image. Next, we describe architecture of our refiner model.

The refiner model takes a rendered image as an input and produces the image with increased realism.  Different component of the proposed loss function are described in Figure \ref{fig:refine_pipeline}. Denote the space of synthetic eye region images as $S$ and the space of realistic eye region images as $R$. The unpaired image-to-image translation procedure involves four different networks:
\begin{itemize}
\item $G$ - a mapper network which learns to map images from $S$ to $R$.
\item $F$ - a mapper network which learns to map images from $R$ to $S$.
\item $D_{S}$ - a discriminator network which learns to detect image from the synthetic domain $S$.
\item $D_{R}$ - a discriminator network which learns to detect image from the realistic domain $R$.
\end{itemize}
To train $G$ to map from the synthetic domain $S$ to the realistic one $R$, we use the least square generative adversarial loss~\cite{mao2016least}
\begin{align}
\mathcal{L}_{LSGAN}(G, D_{S}, S, R) & = \E_{s \thicksim p_{data}(s)} \left[\left( D_{S} (s) - 0.9\right)^2 \right]  \cr & + \E_{r \thicksim p_{data}(r)} \left[ \left(D_{S} (r)\right)^2 \right].
\end{align}
Notice that we used $0.9$ instead of $1.0$ for stabilizing the training process.
We use an equivalent loss function for training the networks $F$ and $D_{R}$.

\begin{figure*}
\begin{center}
%\fbox{\rule{0pt}{4in} \rule{1.0\linewidth}{0pt}}
   \includegraphics[width=1.0\linewidth]{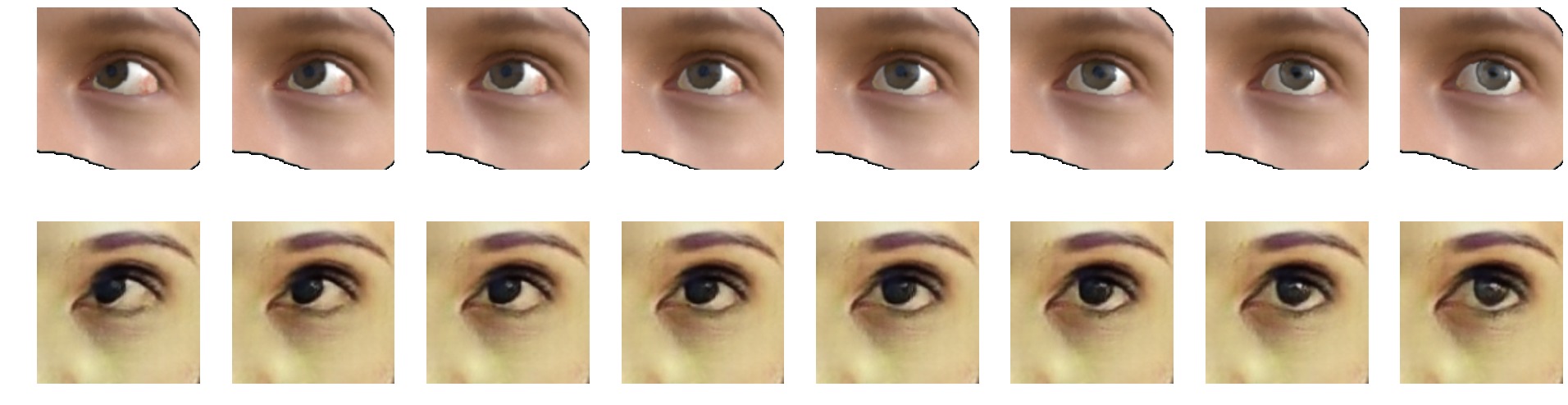}
\end{center}
\vspace{-20pt}
   \caption{Intermediate results.
%Top to bottom: synthetic eye region images drawn from the SynthesEyes model~\cite{wood2014eyetab}, synthetic images with textured mapped from real facial images, refined images without gaze cycle consistency loss, and refined images with gaze cycle consistency loss.
Top row: synthetic eye region images drawn from the SynthesEyes model~\cite{wood2014eyetab} for various gaze yaw angle; bottom -  refined images with gaze cycle consistency loss.}
\vspace{-15pt}
\label{fig:inter_results}
\end{figure*}

Optimizing over the above loss functions trains the networks $G$ and $F$ to map images from one domain into the other.
However, no image feature is enforced to be preserved throughout the mapping as, theoretically, the network can memorize and yield a single image from the target domain and minimize the loss.
To produce the desired maps, additional constraints necessary.
In the original formulation of the method, the following loss function, called cycle-consistency loss, is proposed.
\begin{align}
\mathcal{L}_{cyc} (G,F) & = \E_{s \thicksim p_{data}(s)}\left[  \| F(G(s)) - s \|_1 \right] \cr & + \E_{r \thicksim p_{data}(r)}\left[  \| G(F(r)) - r \|_1 \right].
\end{align}
This loss enforces $F$ and $G$ to encode images and decode images produced by the other network.
Thus, the architecture can be perceived as two interleaved autoencoders.
As demonstrated in Figure \ref{fig:inter_results}, the framework maps synthetic images to realistic ones while preserving the geometric structure of the eye region. We use L1 norm for the reconstruction loss, since it encourages network to produce more sharper looking images, unlike L2 norm, which make the reconstructed images look washed out.

\subsection{Accurate Gaze Direction Annotation}\label{sec:gaze_preserve}
\vspace{-5pt}
Detecting the gaze direction of an eye requires analyzing the pose of the eyeball which can be inferred from the projection of the iris boundaries~\cite{wood2014eyetab}.
Therefore, preserving the geometric structure of the scene is more crucial in the regions around the eyeball than in other parts.
To enforce the translation framework to preserve image features which are more critical for the gaze annotation, we pre-train an additional estimator network, denoted as $E:S\rightarrow \mathbb{R}^3$, to detect the 3D gaze direction on the synthetic images only.
Since the synthetic data has accurate ground truth labels of the optical axis of the eye and the latent space is low dimensional, the architecture of the network $E$ is designed to overfit and to predict the gaze direction with minimal error.
The architectures of the networks are given in the supplementary material of this paper.
Next, we use the network $E$ as an additional constraint on the cyclic translation which maps images from the synthetic domain $S$ to the realistic domain $R$ and back to $S$.
\begin{align}
\mathcal{L}_{gaze-cyc} (G, F) = \E_{s \thicksim p_{data}(s)} \left[ \|E(F(G(s))) - E(s)\|_2^2 \right].
\end{align}
We term this constraint as the {\it gaze cycle consistency loss}.
As demonstrated in Figure \ref{fig:inter_results}, training the framework with the additional loss further preserves the gaze direction in the fake realistic images.
The full objective of the training procedure is
\begin{align}
\mathcal{L}(G, F, D_{S}, D_{R}) &= \mathcal{L}_{LSGAN}(G, D_{S}, S, R) \cr
& + \mathcal{L}_{LSGAN}(F, D_{R}, R, S) \cr
& + \mathcal{L}_{cyc} (G,F) \cr
& + \mathcal{L}_{gaze-cyc} (G, F).
\end{align}

To evaluate the performance of the proposed image synthesis approach, we conducted several experiments.
First, we demonstrate a visual analysis of the robustness of our method to different variations.
Next, we evaluate the proposed technique on the problem of detecting the 2D position on a screen which the subject is looking at.

\section{Experiments}
% Dimensions: size of the image (128x128 - Gazelle, 224x224 - iTracker/Alexnet, 448x448 - full face).
% Report: ME-All, ME-Phone and ME-Tablet.
% Datasets: Real, Synthetic, Refined-AppleGan, Refined-GazelleGAN-{1,2,3}

\begin{table}
\begin{center}
\begin{tabular}{|l|c | c| c |}
\hline
\multirow{2}{*}{Method}  &  \multirow{2}{*}{Training}  & \multicolumn{2}{ |c|}{MeanError(cm)} \\
& data & iPhone & iPhone + iPad \\
\hline
iTracker-R-128 & R & 2.216 & 2.402 \\
iTracker-R-224 & R & 2.182 & 2.348 \\
FullFace-R-128 & R & 3.314 & 3.932 \\
FullFace-R-448 & R & 2.122 & 2.285 \\
SingleEye-R-128 & R & 2.452 & 2.753 \\
%SingleEye-RS-128 & R + S & 2.262 & 2.508 \\
SingleEye-RF-128 & R + F & 2.330 & 2.633 \\
TwoEye-R-128 & R & 2.109 & 2.317 \\
%TwoEye-RS-128 & R + S & 2.045 & 2.270 \\
TwoEye-RF-128 & R + F & 2.009 & 2.220 \\
\hline
\end{tabular}
\end{center}
\vspace{-15pt}
\caption{Prediction error in cm on GazeCapture test set. R here refers to model trained on real data; F for refined; RF for both combined.}
\label{2d-global}
\end{table}

\begin{comment}
\begin{table}
\begin{center}
\begin{tabular}{|l|c | c| c |}
\hline
\multirow{2}{*}{Method}  &  \multirow{2}{*}{Training data}  & \multicolumn{2}{ |c|}{MeanErorr(cm)} \\
& & iPhone 4  & iPhone 6 \\
\hline
iTracker-R-128 & R & 2.170 & 2.263 \\
iTracker-R-224 & R & 2.358 & 2.212 \\
FullFace-R-128 & R & 2.597 & 3.505 \\
FullFace-R-448 & R & 2.143 & 2.143 \\
SingleEye-R-128 & R & 1.937 & 2.526 \\
%SingleEye-RS-128 & R + S & 1.997 & 2.317 \\
SingleEye-RF-128 & R + F & 1.892 & 2.384 \\
TwoEye-R-128 & R & 1.729 & 2.179 \\
%TwoEye-RS-128 & R + S & 1.722 & 2.051 \\
TwoEye-RF-128 & R + F & 1.675 & 2.062 \\
\hline
\end{tabular}
\end{center}
\caption{Prediction error in cm on specific devices.}
\label{2d-per-device}
\end{table}
\end{comment}

\begin{table}
\begin{center}
\begin{tabular}{|l|c| c | c |}
\hline
\multirow{2}{*}{Method}  & \multicolumn{3}{ |c|}{MeanError(cm)} \\
& iPhone 4  & iPhone5 & iPhone 6 \\
\#Training Frames & 12K & 400K & 600K \\
\hline
iTracker-R-128 & 2.170 & 2.133 & 2.263 \\
iTracker-R-224 & 2.358 & 2.121 & 2.212 \\
FullFace-R-128 & 2.597 & 2.987 & 3.505 \\
FullFace-R-448 & 2.143 & 2.082 & 2.143 \\
SingleEye-R-128 & 1.937 & 2.333 & 2.526 \\
%SingleEye-RS-128 & R + S & 1.997 & 2.317 \\
SingleEye-RF-128 & 1.892 & 2.281 & 2.384 \\
TwoEye-R-128 & 1.729 & 2.014 & 2.179 \\
%TwoEye-RS-128 & R + S & 1.722 & 2.051 \\
TwoEye-RF-128 & 1.675 & 1.874 & 2.062 \\
\hline
\end{tabular}
\end{center}
\vspace{-15pt}
\caption{Prediction error in cm on specific devices.}
\vspace{-10pt}
\label{2d-per-device}
\end{table}

% \begin{table}
% \begin{center}
% \begin{tabular}{|l|c | c| c | c|}
% \hline
% Method & MeanErorr(cm) & MeanErorr(cm) on iPhone \\
% \hline
% iTracker-R-128 & 2.402 & 2.216 \\
% iTracker-R-224 & 2.348 & 2.182 \\
% FullFace-R-128 & 3.932 & 3.7 \\
% FullFace-R-448 & 2.285 & 2.122 \\
% SingleEye-R-128 & 2.753 & 2.452 \\
% SingleEye-RS-128 & 2.508 & 2.262 \\
% TwoEye-R-128 & 2.317 & 2.109 \\
% TwoEye-RS-128 & 2.270 & 2.045 \\
% \hline
% \end{tabular}
% \end{center}
% \caption{Prediction error in cm on GazeCapture test set.}
% \end{table}

\begin{figure}
\begin{center}
%\fbox{\rule{0pt}{4in} \rule{1.0\linewidth}{0pt}}
   \includegraphics[width=0.9\linewidth]{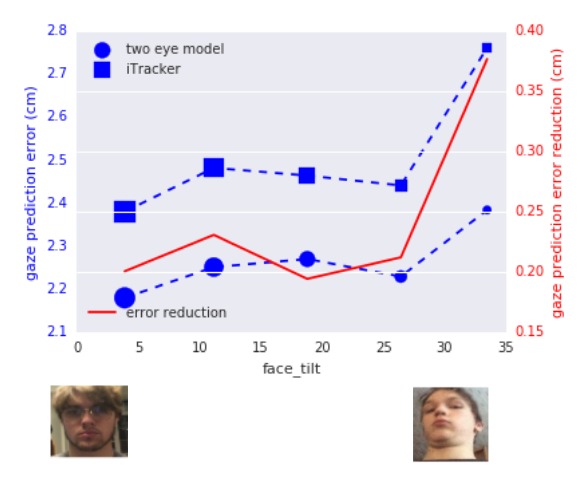}
\end{center}
\vspace{-20pt}
   \caption{Prediction error as a function of head tilt. Marker size represents the number of samples in each bin.}
\vspace{-10pt}
\label{fig:tilt_curve}
%\end{figure}

%\begin{figure}
\begin{center}
%\fbox{\rule{0pt}{4in} \rule{1.0\linewidth}{0pt}}
   \includegraphics[width=0.9\linewidth]{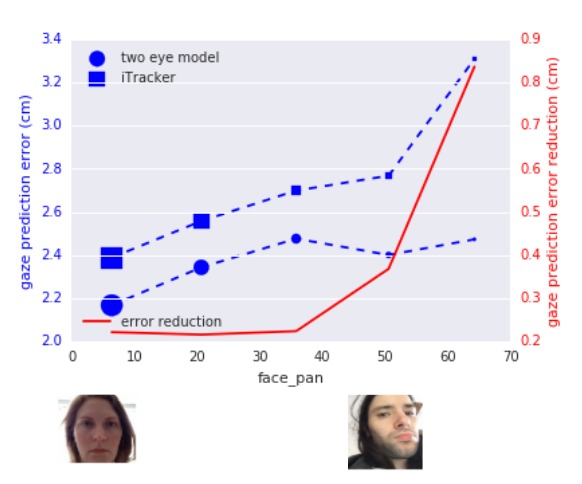}
\end{center}
\vspace{-20pt}
   \caption{Prediction errors as a function of head pan. Marker size represents the number of samples in each bin.}
\vspace{-10pt}
\label{fig:pan_curve}
\end{figure}

\begin{figure}
\begin{center}
%\fbox{\rule{0pt}{4in} \rule{1.0\linewidth}{0pt}}
   \includegraphics[width=0.9\linewidth]{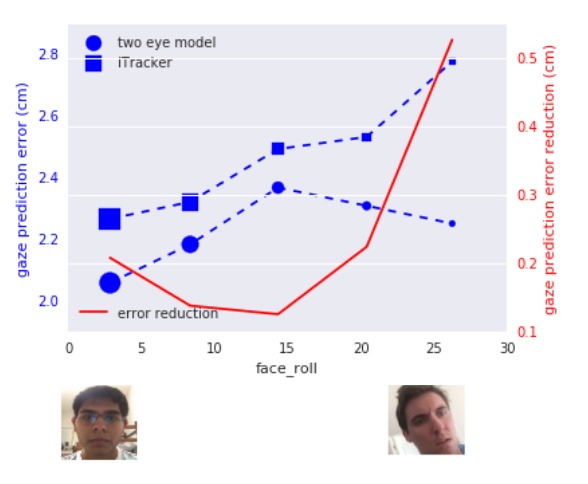}
\end{center}
\vspace{-20pt}
   \caption{Prediction errors as a function of head roll. Marker size represents the number of samples in each bin.}
\vspace{-15pt}
\label{fig:roll_curve}
\end{figure}

\begin{figure}
\begin{center}
%\fbox{\rule{0pt}{4in} \rule{1.0\linewidth}{0pt}}
   \includegraphics[width=0.9\linewidth]{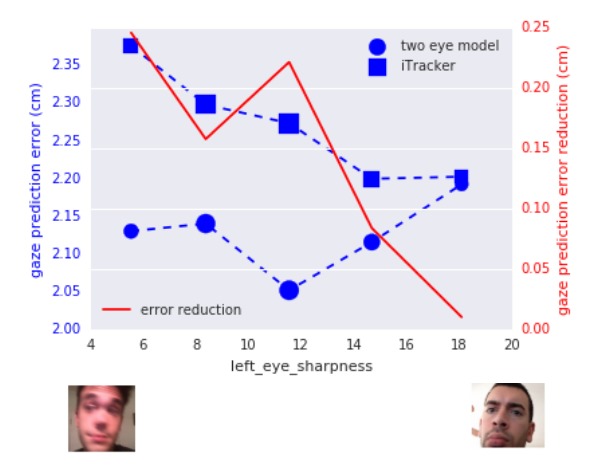}
\end{center}
\vspace{-20pt}
   \caption{Prediction errors as a function of blur/sharpness level. Marker size represents the number of samples in each bin.}
\vspace{-15pt}
\label{fig:blur_curve}
\end{figure}
%\vspace{-15pt}

\begin{figure}
\begin{center}
%\fbox{\rule{0pt}{4in} \rule{1.0\linewidth}{0pt}}
   \includegraphics[width=0.9\linewidth]{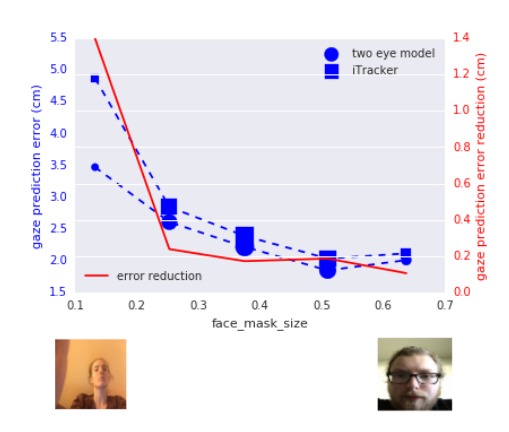}
\end{center}
\vspace{-20pt}
   \caption{Prediction errors as a function of face mask size (proxy for distance). Marker size represents the number of samples in each bin.}
\vspace{-15pt}
\label{fig:distance_curve}
\end{figure}

\subsection{Visual Robustness Evaluation}
To qualitatively analyze the image synthesis method, we evaluate the robustness of the method to different smooth variations.
The proposed setup allows configuring the geometric structure of the eye region, its texture, the camera pose, the gaze direction and other parameters.
First, in Figure \ref{fig:inter_results}, we examine the robustness of the synthesis for the scenario where all the configurations are similar besides the gaze direction.
As demonstrated in the refined images, all the properties are relatively stable, while the gaze changes and is controlled by the synthetic images.

\subsection{2D Gaze Position Estimation}

Next, we evaluate the benefits of the proposed approach against prior work for 2D gaze position estimation. We report the mean prediction error (Euclidean distance) in cms on the screen.

Table \ref{2d-global} reports model comparison on iPhones, as well as iPhones + iPads in the GazeCapture dataset from MIT. As shown in the  table, for iPhones, the iTracker model~\cite{krafka2016eye, kyle2015thesis} trained on real data has a mean error of 2.182cm for 224x224 input images, and 2.216cm for small (128x128) input images \footnote{The model errors from our implementation of iTracker are consistent with the ones reported in ~\cite{kyle2015thesis}, though a bit different from ~\cite{krafka2016eye}.}. The error of full face model~\cite{zhang2016s} is 2.122cm for very large input images of 448x448, but degrades heavily to 3.314cm for small input images of 128x128, since too much information is lost for the small face image, without cropped eye images. In comparison, our two eye model with small input image (128x128; see supplementary for model architecture) has a smaller mean error of 2.109cm when trained with real data only, and improves to 2.009cm when trained with both real and refined data, which is substantially better than prior art (9\% reduction in error compared to iTracker-R-128).

Table \ref{2d-per-device} further slices model performance per device. The proposed two eye model trained with real and synthesized refined images has the lowest error compared to all baseline methods. For example, for a fixed input image size of 128x128, it outpuperforms iTracker model~\cite{krafka2016eye} with reasonable margins (9-22\%), without requiring any additional labeled data from real users. In particular, we get a 22\% reduction in error on iPhone4 (from 2.170 to 1.675cm) where real data is limited ($\sim$12K frames) and a 9-12\% improvement on iPhone6 and iPhone5 respectively, where more real data is available ($\sim$400-600K training frames).
%, i.e., 22\% reduction in error on iPhone4 (from 2.170 to 1.675cm) where real data is limited; 12\% reduction in error on iPhone5 (from 2.133 to 1.874cm), and 9\% reduction in error on iPhone6 (from 2.263cm to 2.062cm). The improvement of our proposed model on iPhone 4 is larger  than iPhone 6 because there are much more data collected from iPhone 6 in GazeCapture dataset.

To test the robustness of our approach to varying head pose, blur and distance to camera, we divide the iPhones dataset into bins of different head pose (absolute values of head tilt, pan and roll), sharpness level  of eye region (inverse of blur), and face\_mask\_size, which is the ratio of the size of face mask over the frame (note: this is a proxy for distance to the camera - higher the ratio, smaller the distance). Figures \ref{fig:tilt_curve}-\ref{fig:distance_curve} show a comparison of the performance of our approach vs.\ iTracker-R-128 model as a function of the above factors. A general trend that is consistent across all these figures is that while the performance of both models degrades with increasing amount of head tilt/pan/roll, blur or distance, we find that iTracker-R-128 model has higher gaze estimation error and deteriorates more rapidly than ours. Thus, we see consistent improvement over prior art (red curve in Figures), with bigger improvements for extreme head pose/blur/distance, demonstrating the robustness of our approach. We find similar results while comparing our TwoEye model with and without refined data (see supplementary), confirming that a key benefit of the proposed framework is in handling cases with limited real data, such as for extreme head pose or blur.

\section{Discussion}
The main contribution of this paper is a stable approach for synthesizing a large dataset of high resolution colorful images with diverse texture and poses, high realism of the extended eye region, and accurate gaze annotations; and using this to demonstrate improved gaze position estimation over state-of-the-art, without collecting additional labeled data from real users.

An interesting observation is that the proposed method completes domains which are not part of the 3D model, such as nose and hair. This work goes beyond refinement of rather small (36x60) grayscale images \cite{shrivastava2017learning} and operates on larger 128x128 colorful images. Unlike \cite{shrivastava2017learning}, the cycle consistency loss makes the GAN training more robust, and does not require more sophisticated training scheme that averages parameter updates among several previous steps. Beyond generating eye region images, this approach can be applied to other application domains which were not investigated here. For example, it is possible to generate images of the entire face from synthetic images, similar to the method proposed in ~\cite{sela2017unrestricted}.

%\textcolor{red}{TODO (dmitry): explicitly address diffs between AppleGAN and our method. e.g., higher res color images: 128x128 vs. 36x60 in former; other qualitative improvements?}

The dataset generated in this paper has a number of desirable characteristics. First, it is diverse in textures and subjects, as the framework maps texture of the skin region around the eye to a 3D model, using a few million textures mined from facial images on the internet. Second, it is diverse in head poses and camera settings, since the 3D model can be used to simulate various conditions. Third, it is realistic as the rendered synthetic images are then mapped to the realistic domain using the novel adversarial training approach. Fourth, it preserves the gaze direction annotations in the synthesized realistic data, via an additional constraint on the gaze cycle consistency loss.

We demonstrate that using the above dataset to train deep learning models on the cropped eye region, yields improvements in gaze estimation performance over prior work, without requiring any additional labeled data from real users. For example, our approach yields an 9\% reduction in error (from 2.216cm in~\cite{krafka2016eye} to 2.009cm) on the iPhones dataset. The improvements are more pronounced when we slice by devices, especially when real data is limited, e.g., 22\% reduction in error on iPhone 4 (from 2.17 to 1.675cm) which has $\sim$12K frames of labeled data. We also observe bigger improvements (e.g., 0.5-1cm reduction in error) for extreme values of head pose, blur and distance (as seen in Figures \ref{fig:tilt_curve}-\ref{fig:distance_curve}), where there is limited real data, demonstrating the robustness of our approach under a large range of operating conditions. This improvement is due to the diversity of our synthesized dataset which allows unlimited training data in various head poses and camera settings, unlike real data where extreme poses are often limited.

It is worth noting that unlike prior work \cite{krafka2016eye, zhang17_cvprw} that require the entire face to be visible, our model operates on the eye region. Yet, due to the additional refined training data, it manages to improve the accuracy of the gaze estimation on the same test dataset. In order to make the gaze estimation from a single eye image possbile, as detailed in the supplementary, we use late fusion of the convolutional network activations (from the eye image) and location of the eye corner landmark in the original image, passed through two fully connected layers FC1(32) and FC2(32).
%The fact that the same model architecture provides an improved accuracy when trained on combination of real and refined data speaks to the potential of our approach to improve gaze estimation accuracy even further.

%achieved without requiring additional labelled data from real users.
%this further by showing that bigger model improvements come when real data is limited.

To ensure a stable translation of images from the synthetic domain to the realistic domain, careful attention should be paid to the similarity between the geometric structures of synthetic and realistic scenes. In particular, we observed that the scale of the iris should be roughly similar to achieve a stable visual translation. In addition, the realistic dataset should be diverse enough for extending the overlap between the distributions of the synthetic and the realistic data.

\textbf{Limitations / Future work:} Although the method produces high quality training images, the translation from the synthetic to the realistic domain can occasionally fail.
We observed it mainly in cases where the synthetic images had extreme poses or peculiar skin textures.
A key improvement to the proposed approach, which was not investigated here is to monitor and distill the generated images with the different networks.
For example, upon mapping a synthetic image to a realistic one and then back to synthetic, we could measure the discrepancy between the gaze of the original one and the reconstructed one.
If the predictions are too far apart, we could drop this example from the generated dataset. We could also use the trained discriminators for measuring the realism and to screen the data samples accordingly. Future work will investigate 3D gaze angle estimation, in addition to the 2D gaze estimation studied here.

\section{Conclusions}
We proposed a framework for generating a large set of diverse, realistic training images for the problem of gaze estimation. The framework maps texture of the skin region around the eye (mined from in-the-wild facial images) to a 3D model, which is used to simulate various head poses and camera settings. The rendered synthetic images are then mapped to the realistic domain using a novel adversarial training approach. This framework results in high resolution, colorful, diverse, realistic looking images of the extended eye region, with accurate gaze annotation.
%The proposed framework allows generating images with higher resolution and more colors compared to the state-of-the-art.
Training deep learning models with these images yields more accurate gaze estimation on real images compared to state-of-the-art for 2D gaze position estimation, without using any additional labeled data from real users. We further demonstrate that this approach provides cross-device generalization of model performance, and yields robust gaze estimation even when real labeled data is limited, such as for extreme head pose/blur/distance.
% \begin{figure*}
% \begin{center}
% \fbox{\rule{0pt}{2in} \rule{.9\linewidth}{0pt}}
% \end{center}
%    \caption{Example of a short caption, which should be centered.}
% \label{fig:short}
% \end{figure*}

% \begin{table}
% \begin{center}
% \begin{tabular}{|l|c|}
% \hline
% Method & Frobnability \\
% \hline\hline
% Theirs & Frumpy \\
% Yours & Frobbly \\
% Ours & Makes one's heart Frob\\
% \hline
% \end{tabular}
% \end{center}
% \caption{Results.   Ours is better.}
% \end{table}

% {\small\begin{verbatim}
%    \usepackage[dvips]{graphicx} ...
%    \includegraphics[width=0.8\linewidth]
%                    {myfile.eps}
% \end{verbatim}
% }

% \begin{figure}[t]
% \begin{center}
%    %\includegraphics[width=0.8\linewidth]{egfigure.eps}
% \end{center}
%    \caption{Example of caption.  It is set in Roman so that mathematics
%    (always set in Roman: $B \sin A = A \sin B$) may be included without an
%    ugly clash.}
% \label{fig:long}
% \label{fig:onecol}
% \end{figure}
{\small
\bibliographystyle{ieee}
\bibliography{references}
}

\newpage
\appendix
%\documentclass[10pt,twocolumn,letterpaper]{article}

% \usepackage{cvpr}
% \usepackage{times}
% \usepackage{epsfig}
% \usepackage{graphicx}
% \usepackage{amsmath}
% \usepackage{amssymb}
% \usepackage{subcaption}
% \usepackage{fixltx2e}
% \usepackage{multirow}
% \usepackage{comment}

% \DeclareMathOperator{\E}{\mathbb{E}}
% Include other packages here, before hyperref.

% If you comment hyperref and then uncomment it, you should delete
% egpaper.aux before re-running latex.  (Or just hit 'q' on the first latex
% run, let it finish, and you should be clear).
% \usepackage[pagebackref=true,breaklinks=true,letterpaper=true,colorlinks,bookmarks=false]{hyperref}
% \begin{document}

%%%%%%%%% TITLE
% \title{Supplementary Material for GazeGAN -- Unpaired Adversarial Image Generation for Gaze Estimation}

% \maketitle
%%%%%%%%%%%%%%%%%%%%%%%%%%%%%%%%%%%%%%%%%%%%%%%%%%%%%%%%%%%%%%%%%%%%%%

\section*{Supplementary Material}
\section{Qualitative Diversity Analysis}
\begin{figure*}
\begin{center}
%\fbox{\rule{0pt}{4in} \rule{1.0\linewidth}{0pt}}
   \includegraphics[width=0.9\linewidth]{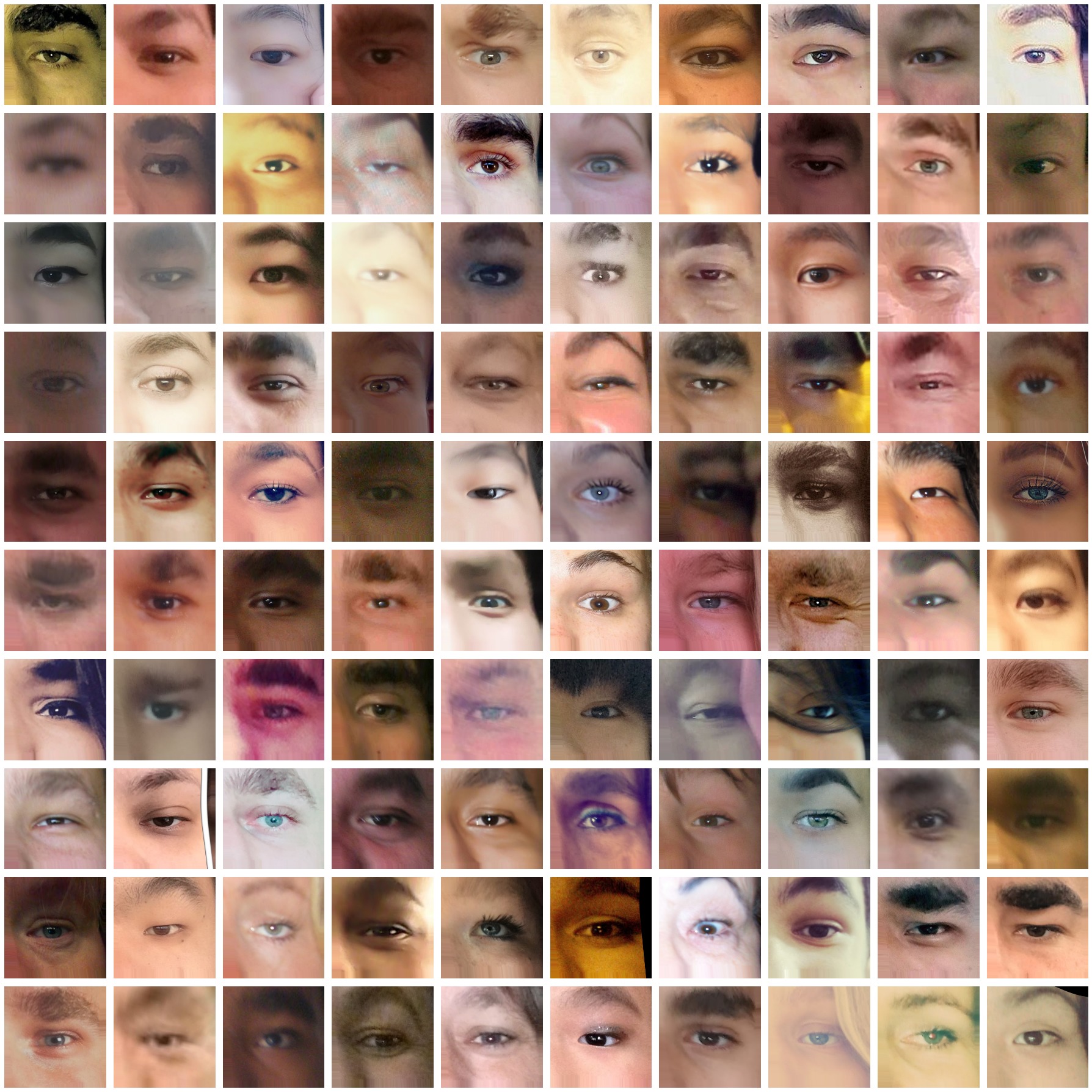}
\end{center}
\vspace{-20pt}
   \caption{Diversity visualization: Different aligned skin textures mined from in-the-wild facial images.}
\vspace{-15pt}
\label{fig:uv_textures}
\end{figure*}

\begin{figure}
\begin{center}
   \includegraphics[width=0.9\linewidth]{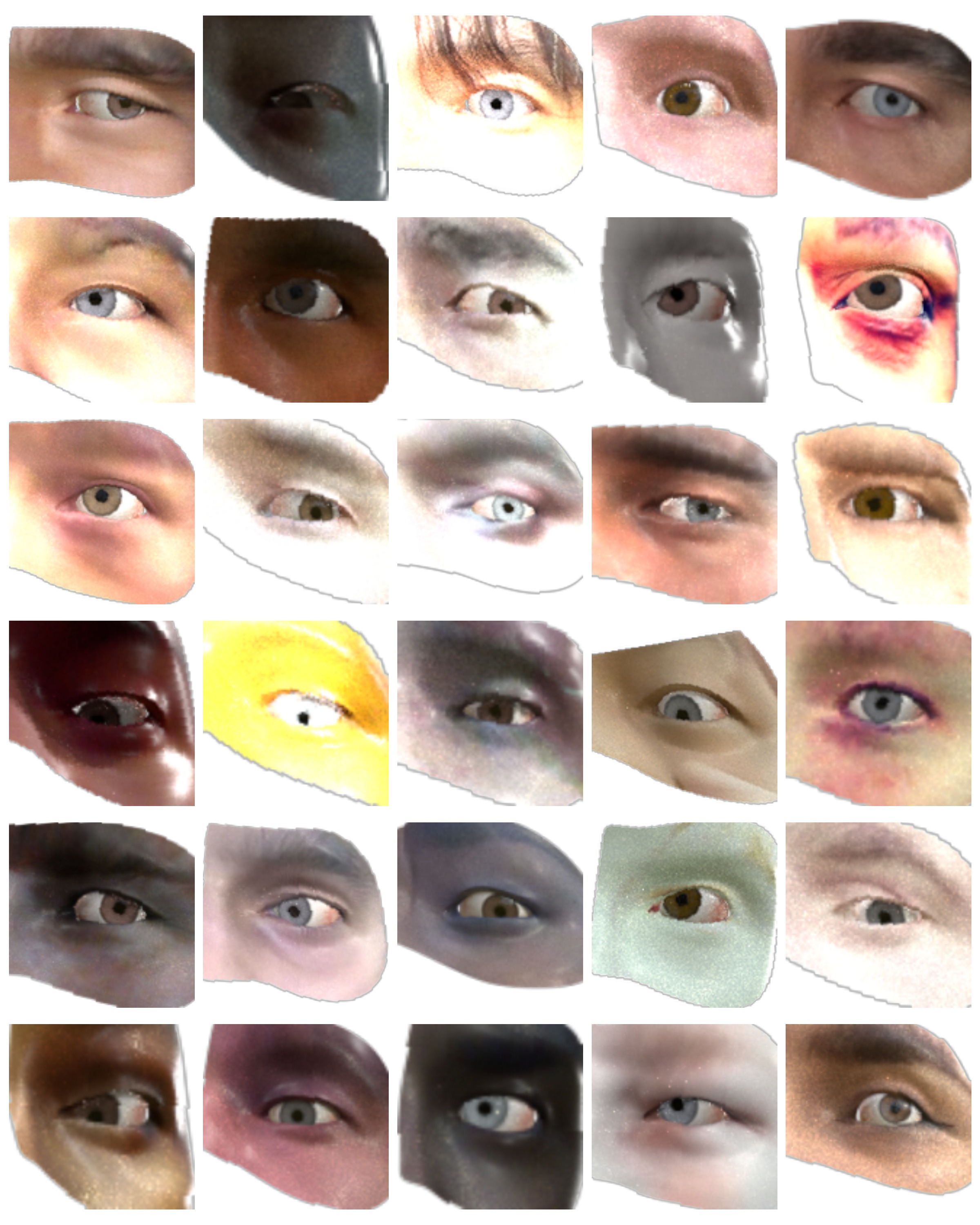}
\end{center}
\vspace{-20pt}
   \caption{Diversity visualization: Synthetic images with textures mapped from in-the-wild facial images.}
\vspace{-15pt}
\label{fig:diverse_syn}
\end{figure}

\begin{figure}
\begin{center}
\includegraphics[width=0.9\linewidth]{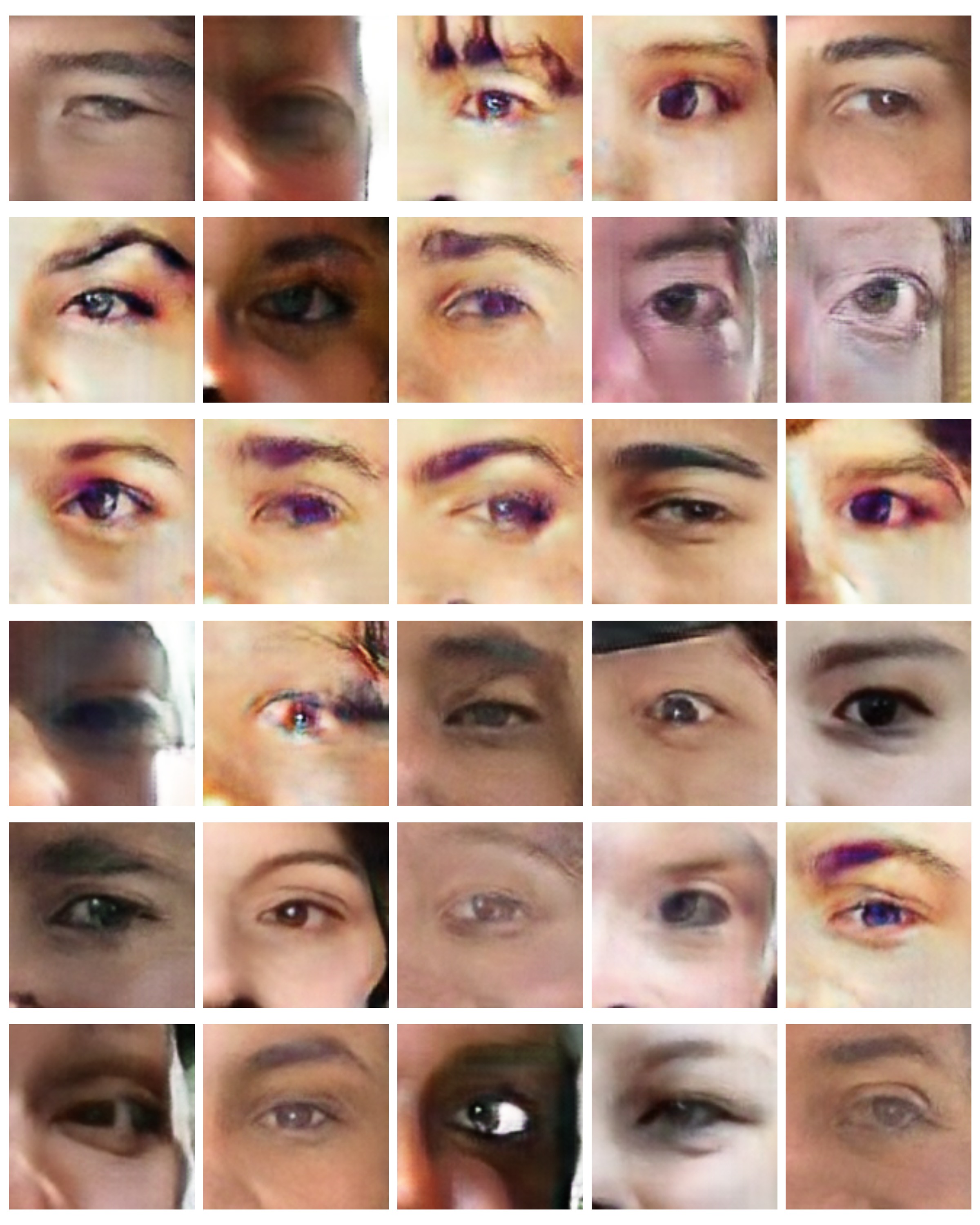}
\end{center}
\vspace{-20pt}
   \caption{Diversity visualization: images produced by the proposed synthetic-to-realistic translation method, corresponding to Figure \ref{fig:diverse_syn}.}
\vspace{-15pt}
\label{fig:diverse_refined}
\end{figure}

\begin{figure*}
\begin{center}
%\fbox{\rule{0pt}{4in} \rule{1.0\linewidth}{0pt}}
   \includegraphics[width=0.9\linewidth]{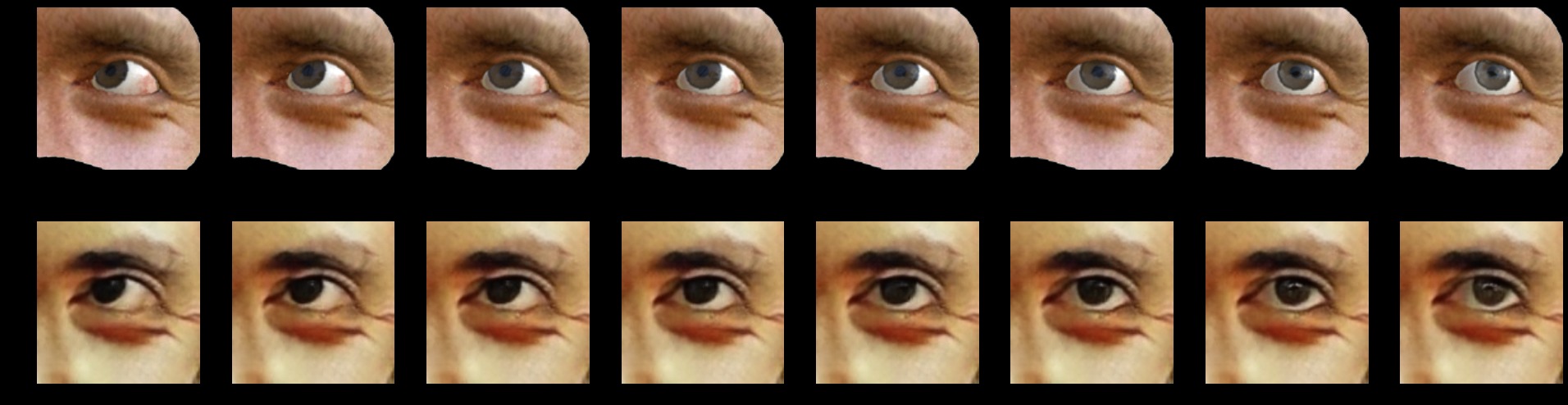}
\end{center}
\vspace{-20pt}
   \caption{Robustness visualization: images produced by the proposed synthetic-to-realistic translation method before (top) and after (bottom) refinement with different gaze direction.}
\vspace{-15pt}
\label{fig:robust_gazedir}
\end{figure*}

A randomly sampled set of the aligned in-the wild facial images to the UV space of the eye region rig is given in Figure \ref{fig:uv_textures}.
With an abundance of these aligned images, we render multiple simulated images of the eye regions with different parameters as in Figure \ref{fig:diverse_syn}.
Finally, we turn these synthetic images into realistic ones while preserving the gaze direction, see Figure \ref{fig:diverse_refined}.
To demonstrate the preservation of the gaze throughout the latter procedure, we present in Figure \ref{fig:robust_gazedir} a sequence of simulated images alongside their refined version, where the only difference between subsequent images is the gaze direction. This shows that the gaze direction in the refined images can be controlled by the simulator.

%%%%%%%%%%%%%%%%%%%%%%%%%%%%%%%%%%%%%%%%%%%%%%%%%%%%%%%%%%%%%%%%%%%%%%
\section{Quantitative Comparison}
\begin{figure}
\begin{center}
%\fbox{\rule{0pt}{4in} \rule{1.0\linewidth}{0pt}}
   \includegraphics[width=0.9\linewidth]{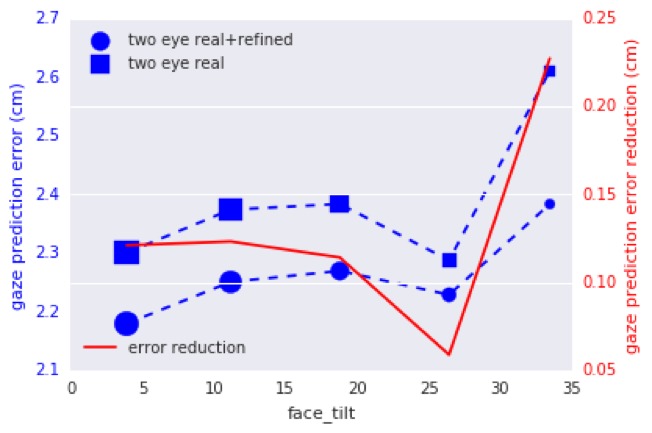}
\end{center}
\vspace{-20pt}
   \caption{Prediction errors as a function of face tilt. Marker size represents the number of samples in each bin.}
\vspace{-15pt}
\label{fig:face_tilt}
\end{figure}

\begin{figure}
\begin{center}
%\fbox{\rule{0pt}{4in} \rule{1.0\linewidth}{0pt}}
   \includegraphics[width=0.9\linewidth]{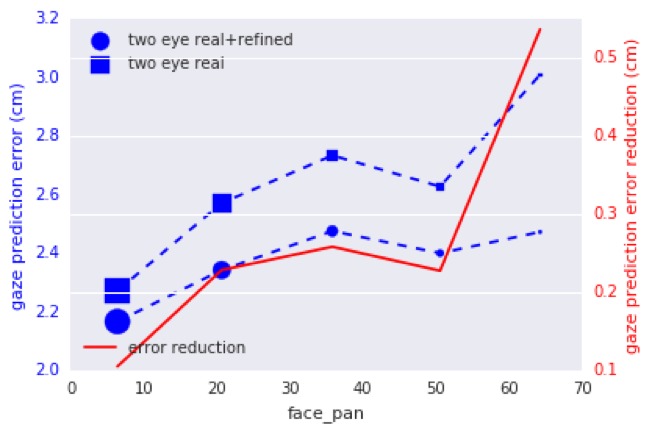}
\end{center}
\vspace{-20pt}
   \caption{Prediction errors as a function of face pan. Marker size represents the number of samples in each bin.}
\vspace{-15pt}
\label{fig:face_pan}
\end{figure}

\begin{figure}
\begin{center}
%\fbox{\rule{0pt}{4in} \rule{1.0\linewidth}{0pt}}
   \includegraphics[width=0.9\linewidth]{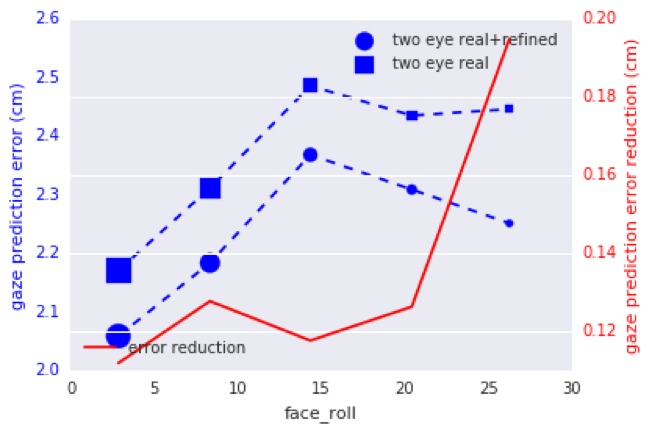}
\end{center}
\vspace{-20pt}
   \caption{Prediction errors as a function of face roll. Marker size represents the number of samples in each bin.}
\vspace{-15pt}
\label{fig:face_roll}
\end{figure}

\begin{figure}
\begin{center}
%\fbox{\rule{0pt}{4in} \rule{1.0\linewidth}{0pt}}
   \includegraphics[width=0.9\linewidth]{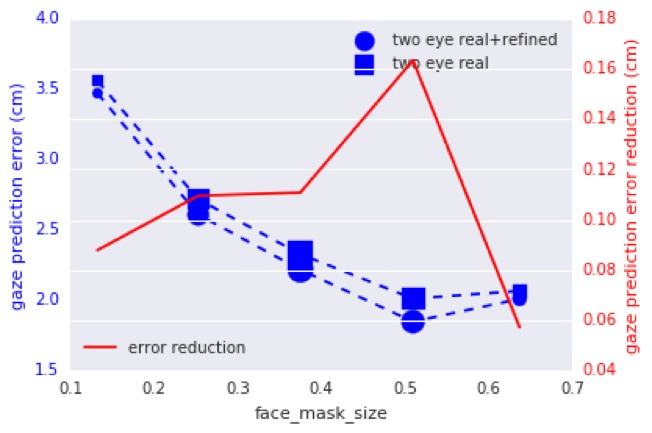}
\end{center}
\vspace{-20pt}
   \caption{Prediction errors as a function of facial mask size. Marker size represents the number of samples in each bin.}
\vspace{-15pt}
\label{fig:face_mask}
\end{figure}

\begin{figure}
\begin{center}
%\fbox{\rule{0pt}{4in} \rule{1.0\linewidth}{0pt}}
   \includegraphics[width=0.9\linewidth]{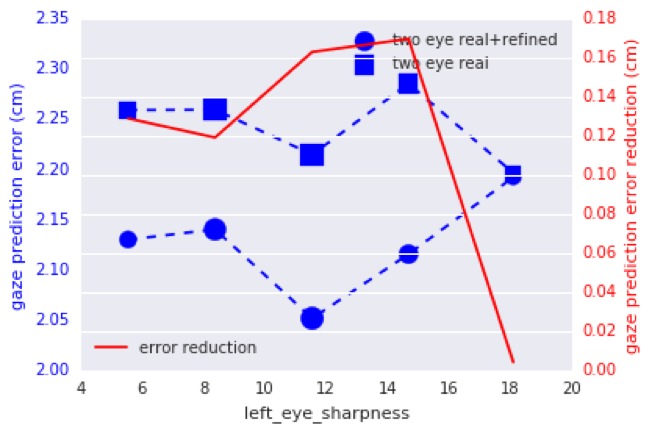}
\end{center}
\vspace{-20pt}
   \caption{Prediction errors as a function of image sharpness. Marker size represents the number of samples in each bin.}
\vspace{-15pt}
\label{fig:sharpness}
\end{figure}

To measure the improvement in performance with the proposed synthesized data versus using real data only, we evaluated the gaze estimation error under different variations in the real data.
First, in Figures \ref{fig:face_tilt}, \ref{fig:face_pan} and \ref{fig:face_roll}, the accuracy of the gaze estimation is measured under different face poses as a function of the facial tilt, pan and roll angles, respectively. Notice the error reduction rate with the proposed data.
Next, in Figure \ref{fig:face_mask}, we evaluate the reduction of the error as a function of the ratio between the area of the facial mask and the entire image. As can be expected, the error decrease as the face is closer to the camera. Yet in this comparison we observed only mild improvement with the refined data.
Finally, we estimate in Figure \ref{fig:sharpness} the error as a function of sharpness of input images around the left eye. Notice that the refined images increased the performance on test images which are more blurred while for sharp images the contribution is negligible.

%%%%%%%%%%%%%%%%%%%%%%%%%%%%%%%%%%%%%%%%%%%%%%%%%%%%%%%%%%%%%%%%%%%%%%%%%%%

\section{Network Architectures and Parameters}
\subsection{Image-to-Image Translators}
The proposed image synthesis method involves five instances of three different networks,
a refiner image-to-image mapper, a discriminator network and a pre-trained 3D gaze estimator.
The architecture of each of these networks are given in Tables \ref{table:arch_refiner}, \ref{table:arch_discriminator}, and \ref{table:arch_pretrained_gaze}, respectively.
For training the image synthesis method we used ADAM algorithm ~\cite{kingma2014adam} with learning rate of 0.00001, $\beta_1 = 0.1$ and $\beta_2 = 0.99$.

\begin{table}
\begin{center}
\begin{tabular}{ c | c }
\hline
\hline
Type / Stride & Filter Shape  \\
\hline % inserts single horizontal line
Conv / s1 & $7\times7\times3\times32$ \\
\hline
Instance Normalization &  \\
\hline
ReLU &  \\
\hline
Conv / s2 & $3\times3\times32\times64$ \\
\hline
Instance Normalization &  \\
\hline
ReLU &  \\
\hline
Conv / s2 & $3\times3\times64\times128$ \\
\hline
Instance Normalization &  \\
\hline
ReLU &  \\
\hline
 $6\times$Residual Blocks & $3\times3\times128\times128$ \\
\hline
Conv / s0.5 & $3\times3\times128\times64$ \\
\hline
Instance Normalization &  \\
\hline
ReLU &  \\
\hline
Conv / s0.5 & $3\times3\times64\times32$ \\
\hline
Instance Normalization &  \\
\hline
ReLU &  \\
\hline
Conv / s1 & $7\times7\times32\times3$ \\
\hline
Instance Normalization &  \\
\hline
ReLU &  \\
\hline
\end{tabular}
\end{center}
\caption{Architecture of the refiner networks.}
\label{table:arch_refiner}
\end{table}

\begin{table}
\begin{center}\begin{tabular}{ c | c }
\hline
\hline
Type / Stride & Filter Shape  \\
\hline
Conv / s2 & $4\times4\times3\times64$ \\
\hline
LeakyReLU - slope 0.2 &  \\
\hline
Conv / s2 & $4\times4\times64\times128$ \\
\hline
Instance Normalization &  \\
\hline
LeakyReLU - slope 0.2 &  \\
\hline
Conv / s2 & $4\times4\times128\times256$ \\
\hline
Instance Normalization &  \\
\hline
LeakyReLU - slope 0.2 &  \\
\hline
Conv / s2 & $4\times4\times256\times512$ \\
\hline
Instance Normalization &  \\
\hline
LeakyReLU - slope 0.2 &  \\
\hline
Conv / s1& $?\times?\times512\times1$
\end{tabular}
\end{center}
\caption{Architecture of the discriminator networks. The last layer computes a convolution of size dependent upon the input size, such that the output will be of dimension 1.}
\label{table:arch_discriminator}
\end{table}

\begin{table}
\begin{center}
\begin{tabular}{ c | c }
\hline
\hline
Type / Stride & Filter Shape  \\
\hline % inserts single horizontal line
Conv / s2 & $3\times3\times3\times32$ \\
\hline
Conv dw / s1& $3\times3\times32$ dw \\
\hline
Conv / s1& $1\times1\times32\times64$ \\
\hline
Conv dw / s2& $3\times3\times64$ dw \\
\hline
Conv / s1& $1\times1\times64\times128$ \\
\hline
Conv dw / s1& $3\times3\times128$ dw \\
\hline
Conv / s1& $1\times1\times128\times128$ \\
\hline
Conv dw / s2& $3\times3\times128$ dw \\
\hline
Conv / s1& $1\times1\times128\times256$ \\
\hline
Conv dw / s1& $3\times3\times256$ dw\\
\hline
Conv / s1& $1\times1\times256\times256$ \\
\hline
Conv dw / s2& $3\times3\times256$ dw \\
\hline
Conv / s1& $1\times1\times256\times512$ \\
\hline
\multirow{2}{*}{$5\times$} Conv dw / s1& $3\times3\times512$ dw \\
\hspace{.53cm}Conv / s1\\
\hline
Conv dw / s2& $3\times3\times512$ dw \\
\hline
Conv / s1& $1\times1\times512\times1024$ \\
\hline
Conv dw / s2& $3\times3\times1024$ dw \\
\hline
Conv / s1& $1\times1\times1024\times1024$ \\
\hline
Avg Pool / s1& Pool $7\times7$ \\
\hline
FC / s1 & $1024 \times 1000$ \\
\hline
Softmax / s1 & Classifier\\
\hline %inserts single line

\end{tabular}
\end{center}
\caption{Architecture of the pre-trained 3D gaze estimator network. "dw" refers to - depth-wise layers (see Mobilenet~\cite{howard2017mobilenets} for more information).}
\label{table:arch_pretrained_gaze}
\end{table}

\subsection{2D Gaze Estimator - Single Eye}
\label{gaze-estimator-single-eye}

\begin{figure*}
\begin{center}
\includegraphics[width=0.9\linewidth]{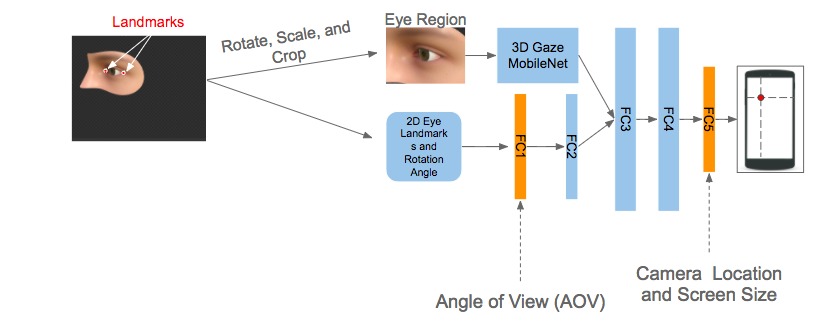}
\end{center}
\vspace{-20pt} \caption{Architecture of the 2D gaze estimator.}
\vspace{-15pt}
\label{fig:gaze_estimator}
\end{figure*}

To evaluate the performance of the proposed image synthesis approach, we trained a 2D gaze estimator from a single eye image with the architecture shown on Figure \ref{fig:gaze_estimator}. Our model takes eye region image, an angle used to rotate the image to in order to make it horizontal, and location of eye corners in the original image. The core of the model is the MobileNet architecture \cite{howard2017mobilenets} that is described in Table \ref{table:arch_pretrained_gaze}. Since the gaze location distribution differs for different mobile device models and screen orientations, we include a small set of device specific parameters to let model better fit the data. Specifically, we add device specific shift and scale parameters (for both x and y) for the landmark input of the model, and the output of the model. These parameters are learned during the gaze estimator model training. We use the following sizes for the fully connected layers (shown in Figure \ref{fig:gaze_estimator}): \textsc{FC1(100)}, \textsc{FC2(32)}, \textsc{FC3(32)}, \textsc{FC4(64)}, \textsc{FC5(64)}. To train the network, we used the Adam optimization algorithm~\cite{kingma2014adam} with learning rate of 0.01, $\beta_1 = 0.1$ and $\beta_2 = 0.99$.

\subsection{2D Gaze Estimator - Two Eyes}
\label{gaze-estimator-two-eyes}
Next, we extend our approach for binocular gaze estimation by incorporating image of the left eye into the model. Specifically, we train a separate network that performs gaze estimation based on the left eye image, and then we average the predictions. Once could certainly improve upon this to train more expressive function that combines predictions or pan-terminal activations of the two eye models. We train this network in the same regime as the monocular model --- using the Adam optimization algorithm~\cite{kingma2014adam} with learning rate of 0.01, $\beta_1 = 0.1$ and $\beta_2 = 0.99$.

{\small
\bibliographystyle{ieee}
\bibliography{references}
}

%%%%%%%%%% Merge with supplemental materials %%%%%%%%%%
\end{document}